\documentclass[10pt,twocolumn,letterpaper]{article}

\usepackage[pagenumbers]{cvpr} 

\usepackage[dvipsnames]{xcolor}

\definecolor{cvprblue}{rgb}{0.21,0.49,0.74}
\usepackage[pagebackref,breaklinks,colorlinks,citecolor=cvprblue]{hyperref}

\usepackage[super]{nth}
\usepackage{mathtools}
\usepackage{hyperref}
\usepackage{gensymb}
\usepackage{amsmath}
\usepackage[ruled]{algorithm2e}
\usepackage{breakcites}
\usepackage{booktabs}
\usepackage{multirow}
\usepackage{adjustbox,lipsum}
\usepackage[T1]{fontenc}
\usepackage[accsupp]{axessibility} 

\usepackage{listings}
\usepackage{bm}
\renewcommand{\thefootnote}{\fnsymbol{footnote}}

\def\paperID{9965} 
\def\confName{CVPR}
\def\confYear{2024}

\title{Jointly Training and Pruning CNNs via Learnable Agent Guidance and Alignment}
\author{\stepcounter{footnote}Alireza Ganjdanesh \textsuperscript{\rm 1}$^{*}$
\and 
Shangqian Gao \textsuperscript{\rm 2}$^{*}$
\and
Heng Huang \textsuperscript{\rm 1}\thanks{This work was partially supported by NSF IIS 2347592, 2347604, 2348159, 2348169, DBI 2405416, CCF 2348306, CNS 2347617.} 
\and
\textsuperscript{\rm 1} Department of Computer Science, University of Maryland College Park
\and
\textsuperscript{\rm 2} Department of Electrical and Computer Engineering, University of Pittsburgh\\
{\tt\small \{aliganj, heng\}@umd.edu, shg84@pitt.edu}
}
\begin{document}
\maketitle
\def\thefootnote{*}\footnotetext{These authors contributed equally to this work.} 
\begin{abstract}
Structural model pruning is a prominent approach used for reducing the computational cost of Convolutional Neural Networks (CNNs) before their deployment on resource-constrained devices.~Yet, the majority of proposed ideas require a pretrained model before pruning, which is costly to secure.~In this paper, we propose a novel structural pruning approach to jointly learn the weights and structurally prune architectures of CNN models.~The core element of our method is a Reinforcement Learning (RL) agent whose actions determine the pruning ratios of the CNN model's layers, and the resulting model's accuracy serves as its reward. We conduct the joint training and pruning by iteratively training the model's weights and the agent's policy, and we regularize the model's weights to align with the selected structure by the agent. The evolving model's weights result in a dynamic reward function for the agent, which prevents using prominent episodic RL methods with stationary environment assumption for our purpose. We address this challenge by designing a mechanism to model the complex changing dynamics of the reward function and provide a representation of it to the RL agent. To do so, we take a learnable embedding for each training epoch and employ a recurrent model to calculate a representation of the changing environment. We train the recurrent model and embeddings using a decoder model to reconstruct observed rewards.~Such a design empowers our agent to effectively leverage episodic observations along with the environment representations to learn a proper policy to determine performant sub-networks of the CNN model.~Our extensive experiments on CIFAR-10 and ImageNet using ResNets and MobileNets demonstrate the effectiveness of our method.

\end{abstract}

\section{Introduction} \label{sec:intro}
Convolutional Neural Networks (CNNs) have enabled unprecedented achievements in the last decade in different domains~\cite{he2016deep,redmon2016you,ren2015faster,wang2022yolov7}.~They have shown a trend for better performance when benefiting from deeper and wider architectures, larger dataset sizes, and longer training times with modern hardware~\cite{dollar2021fast,liu2023more,liu2022convnet,woo2023convnext}.~Despite their accomplishments, the tremendous memory and computational requirements of CNNs prohibit deploying them on edge devices with limited battery and compute resources, making CNN compression a crucial step before their deployment.~The goal is to reduce the size and computational burden of CNNs while preserving their performance.~Model pruning (removing weights~\cite{han2015learning} or structures~\cite{li2016pruning} like channels and layers), weight quantization~\cite{rastegari2016xnor}, knowledge distillation~\cite{hinton2015distilling}, Neural Architecture Search (NAS)~\cite{zoph2017neural,howard2019searching}, and lightweight architecture designs~\cite{howard2017mobilenets,sandler2018mobilenetv2} are common categories of ideas for CNN compression.

Structural pruning which removes redundant channels of a CNN is the main focus of this paper.~It is more practically plausible than weight pruning as it can effectively reduce the inference cost of a model on established hardware like GPUs without requiring special libraries~\cite{han2016eie} or post-processing steps.~Further, it demands far less design efforts than NAS~\cite{han2020model} and architecture design methods~\cite{tan2019efficientnet,ma2018shufflenet}.~The proposed structural pruning methods determine the importance of each channel using metrics such as resource loss~\cite{gao2020discrete}, norm~\cite{li2016pruning}, and accuracy~\cite{liu2019metapruning} and prune a model with techniques like greedy search~\cite{ye2020good} as well as evolutionary algorithms~\cite{chin2020towards}.~Thanks to the advances of Reinforcement Learning (RL) methods in complex decision making tasks~\cite{silver2016mastering,berner2019dota,fawzi2022discovering}, leveraging RL methods to determine proper sub-networks of a CNN given a desired budget has been explored in recent years~\cite{he2018amc,ashok2018nn,wu2018blockdrop,zhang2020learning,huang2018learning,yu2022topology}.~AMC~\cite{he2018amc} trains a DDPG agent~\cite{DBLP:journals/corr/LillicrapHPHETS15} to prune layers of a pretrained CNN.~LFP~\cite{huang2018learning} trains an agent to get weights of a CNN's filters and determine keeping or pruning them.~N2N~\cite{ashok2018nn} uses two agents to perform layer removal and layer shrinkage respectively.~Finally, GNNRL~\cite{yu2022topology} utilizes graph neural networks to identify CNN topologies and employ RL to find a proper compression policy.~Despite their promising results, these models require a pretrained CNN model for training their RL agent for pruning as the prominent RL algorithms, like DDPG~\cite{DBLP:journals/corr/LillicrapHPHETS15} used in AMC~\cite{he2018amc}, cannot perform well in dynamic environments~\cite{khetarpal2022towards} if one trains model's weights along with the agent's policy.

We propose a novel model pruning method to jointly learn a CNN's weights and structurally prune its architecture using an RL agent. As training the weights and pruning them cannot happen simultaneously, we apply a soft regularization term to the model's weights during training to align with the sub-structure chosen by the best agent's policy. We iteratively train the model's weights for one epoch and perform several RL trajectories observations on the most recent model to update the policy of the RL agent.~We design our RL agent so that in each of its episodic trajectories, its actions determine the pruning ratio for each layer of the model.~After pruning all layers, we take the resulting model's accuracy as our agent's reward.~However, as the model's weights get updated in each epoch, the reward function of the RL agent changes dynamically between epochs.~Accordingly, the training episodes of our agent are drawn from a non-stationary distribution.~Therefore, one cannot simply employ prominent RL methods like DDPG~\cite{DBLP:journals/corr/LillicrapHPHETS15} and Soft Actor-Critic (SAC)~\cite{haarnoja2018soft} in our framework to prune the model because their core assumption which is the environment being stationary~\cite{khetarpal2022towards} is not fulfilled.~To overcome this challenge, we design a mechanism to model the evolving dynamics of the agent's environment.~We take an embedding for each epoch of the training and employ a recurrent model to determine a representation of the current state of the environment for the agent given the embeddings of the epochs so far.~We train the recurrent model along with a decoder model in an unsupervised fashion to reconstruct the reward values observed in the agent's trajectories.~Finally, we augment each episodic trajectory of the agent with the representations of the state of the environment (provided by the recurrent model) at the time of the trajectory.~By doing so, our RL agent has access to all information regarding the dynamic environment, and we employ SAC~\cite{haarnoja2018soft} to train the agent using the augmented trajectories.~In addition, our soft regularization scheme for alignment of weights and the selected structure by the agent enables our pruned model to readily recover its performance in fine-tuning.~We summarize our contributions as follows:

\begin{itemize}
	\item We propose a novel channel pruning method that jointly learns the weights and prunes the architecture of a CNN model using an RL agent. In contrast with previous methods using RL for pruning, our method does not need a pretrained model before pruning.
        \item We perform joint training and pruning by  iteratively training the model's weights and the agent's policy.~We utilize a soft regularization technique to the model's weights during training, encouraging them to align with the structure determined by the agent.~By doing so, our method can identify a high-performing base model with weights that closely match the structure selected by the agent. Consequently, the pruned model can readily recover its high performance in fine-tuning.

        \item We design a mechanism to model the dynamics of our evolving pruning environment.~To do so, we use a recurrent model that provides a representation of the state of the environment to the agent. We augment the trajectories observed by the agent using the provided representations to train the agent.
	
\end{itemize}

\begin{figure*}[t!]
  \centering
  \includegraphics[scale=0.21]{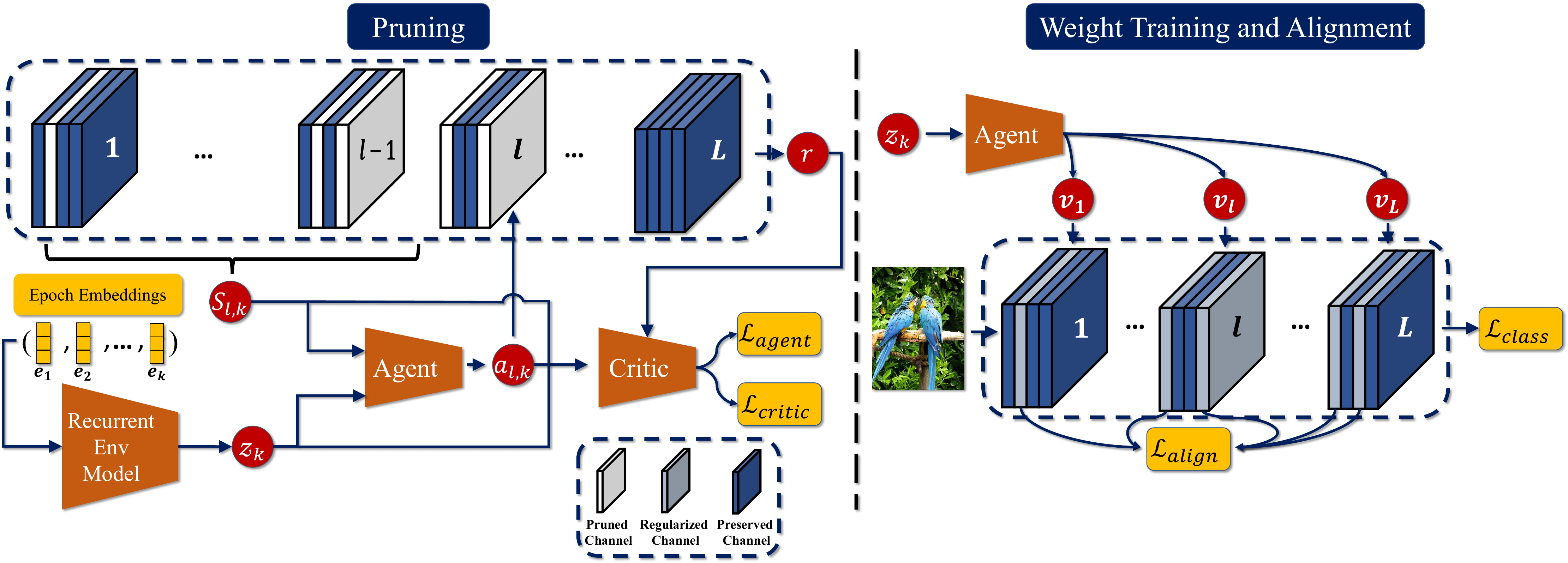}
  \caption{\textbf{Overview of our method}.~We jointly train and prune a CNN model using an RL agent by iteratively training the agent's policy and model's weights.~In each iteration, we train the model's weights for one epoch and perform several episodic observations of the agent.~\textbf{Left:} Each action of our agent prunes one layer of the model, and the procedure of pruning the $l$-th layer is shown.~The agent's actions on the previous layers and the remaining layers' FLOPs determine its state, and we take the resulting model's accuracy as its reward~(Sec.~\ref{iterative-training}).~As the model's weights change between iterations, the reward function also changes accordingly.~Thus, we map each epoch to an embedding and employ a recurrent model to provide a state of the environment $z$ to the agent.~(Sec.~\ref{dynamic-rewards})~\textbf{Right:} Given a sub-network selected by the agent, we train the model's weights while softly regularizing them to align with the selected structure~(Sec.~\ref{soft-reg}).}
  \label{scheme}
\vspace{-10pt}
\end{figure*}

\section{Related Work}
\noindent\textbf{Model Pruning}: Model compression ideas~\cite{ghimire2022survey,marino2023deep} can be categorized as structural pruning~\cite{li2016pruning,molchanov2019importance,peng2019collaborative,sui2021chip,wen2016learning,yin2022hodec,zhuang2018discrimination,gao2022disentangled,gao2022learning,gao2023structural,ganjdanesh2023compressing}, weight quantization~\cite{rastegari2016xnor,wang2022learnable,chen2015compressing}, weight pruning~\cite{han2015learning,yu2022combinatorial,frankle2018the}, Neural Architecture Search (NAS)~\cite{white2023neural,zoph2017neural,ganjdanesh2023effconv}, knowledge distillation~\cite{DBLP:journals/ijcv/GouYMT21}, and lightweight architecture design~\cite{tan2019efficientnet,han2020model}.~Structural pruning focusing on removing redundant channels~(filters) of a CNN is the direction related to this paper.~The proposed methods have approached this problem from various directions like pruning filters with smaller norms~\cite{li2016pruning}, applying group regularization~\cite{he2017channel} like LASSO~\cite{tibshirani1996regression} during training, ranking filters' importance using low-rank decomposition~\cite{lin2020hrank}, estimating influence of a filter on loss~\cite{molchanov2017pruning} using the Taylor decomposition, and meta-learning~\cite{liu2019metapruning}.~Recently, several ideas have employed Reinforcement Learning (RL) for pruning~\cite{he2018amc,ashok2018nn,wu2018blockdrop,zhang2020learning,huang2018learning,yu2022topology}.~LSEDN~\cite{zhang2020learning} trains an RL agent to perform layer-wise pruning on DenseNet~\cite{huang2017densely}.~N2N~\cite{ashok2018nn} proposed a two-stage process that employs two recurrent RL agents in which one agent removes layers of a pretrained CNN, and then, the other agent shrinks each remaining layer.~LFP~\cite{huang2018learning} introduces `try-and-learn' scheme in which RL agents learn to take layers' weights and predict binary masks for pruning or preserving layers' filters.~AMC~\cite{he2018amc} takes a pretrained CNN and trains a DDPG~\cite{DBLP:journals/corr/LillicrapHPHETS15} to prune its convolution layers.~Yet, these models can only prune pretrained models and cannot jointly train and prune the models.~The main reason is that they employ off-the-shelf RL methods~\cite{DBLP:journals/corr/LillicrapHPHETS15,haarnoja2018soft} that assume stationary training environments, but if the model's weights change during pruning, the environment will not be stationary.~We develop a novel channel pruning method that jointly learns a CNN's weights and prunes its architecture using an RL agent in an iterative manner.~We design a procedure to model the changing dynamics of the reward function of our RL agent using a recurrent model that provides a representation of the current state of the environment.~In addition, we regularize the model's weights to align with the structure determined by the agent.~Thanks to such designs, our method can train a base model and select a perfromant sub-network of it that can easily recover its performance in fine-tuning.

\noindent\textbf{Reinforcement Learning}: RL methods~\cite{9904958, huang2022bregman} have achieved outstanding results in complex tasks~\cite{silver2016mastering,berner2019dota,fawzi2022discovering} using techniques like Q-learning~\cite{watkins1992q} and policy optimization~\cite{sutton1999policy}.~Yet, it has been shown that they cannot generalize to new variations of their primary task~\cite{bengio2020interference,khetarpal2022towards}.~Continual RL methods~\cite{khetarpal2022towards} are related to this paper as our agent's environment is non-stationary.~The proposed ideas address different non-stationarity conditions.~For example, multi-task learning~\cite{yu2020gradient} and meta-learning methods~\cite{finn2017model,duan2017rl,wang2016learning} assume that sequential tasks presented to the agent have an unknown stationary distribution.~Curriculum learning ideas~\cite{schmidhuber2013powerplay,ijcai2020p671,chen2021variational} aim to learn the agent's own curriculum in single or multi-agent dynamic environments. Finally, a group of methods~\cite{cheung2020reinforcement,domingues2021kernel,touati2020efficient} make assumptions on the variation budget of the reward function to improve learning of the agent.~We refer to~\cite{khetarpal2022towards} for a comprehensive review on continual RL methods.~Different from the mentioned ideas, we employ a recurrent model to model the changing dynamics of the environment of our agent and augment its observations with it to enable using episodic RL methods to train our agent.
\section{Method}
We propose a new channel pruning method for CNNs to jointly learn the weights of a CNN model and prune its filters using an RL agent.~To do so, we iteratively train the model's weights and the agent's policy.~We design the agent such that its actions determine the compression rate of the model's layers, and we take the accuracy of the pruned model as the reward function for the agent.~Nevertheless, as we update the model's weights iteratively, the reward associated with a certain action changes in consecutive iterations, which results in a non-stationary environment for the agent and prevents using episodic RL methods for our purpose as they assume the environment is stationary. We develop a mechanism in which a recurrent network models the changing dynamics of the reward function and provides a representation of the changing state of the environment to the agent to alleviate this challenge.~Finally, as we cannot both train and prune the weights simultaneously during training, we propose a regularization to align the model's weights with the selected sub-network by the agent.~Fig.~\ref{scheme} shows the overall scheme of our method.

\subsection{Notations}
We denote the number of layers in a CNN with $L$ and the weights of its $l$-th convolution layer with $\mathcal{W}_l\in \mathcal{R}^{C_{l+1}\times C_l\times W_l\times H_l}$.~$C_{l+1}$, $C_l$ represent the number of output and input channels of the layer, and $W_l$ as well as $H_l$ are the spatial dimensions of its kernel. We show the stride of the $l$-th layer with $stride_l$, and $\text{FLOPs}[l]$ is its FLOPs value.~Finally, we show the floor function with $\lfloor \cdot \rfloor$

\subsection{Iterative Weight Training and Compression}\label{iterative-training}
We iteratively train a CNN's weights and optimize the policy of an RL agent to jointly train and prune it.~Specifically, in each iteration, we first train the model's weights for one epoch on our training dataset while the agent is fixed.~Then, we keep the model's weights frozen and our agent observes several episodic trajectories by performing its actions on the CNN's layers.~To employ an RL agent to prune our model, we need to define three main components for the agent: States that it visits in the environment, the actions that it can perform, and the reward function given states and actions.~We describe our choices for each one in the following, and we denote the desired FLOPs budget for the pruned model with $\text{FLOPs}_{desire}$.

\noindent\textbf{States of the Agent:} We design our RL agent to perform actions that determine the pruning rate for consecutive layers of the CNN model.~Thus, the agent's states depend on the index of the layer that the agent is currently pruning; the layer's characteristics like its kernel size and FLOPs; and the number of FLOPs that the agent has already pruned as well as the amount it has to prune from the remaining layers.~Formally, given that the agent is currently pruning the $l$-th layer, we define the state of the environment as follows:

\vspace{-10pt}
\begin{equation}\label{state-define}
	\begin{aligned}
		\small S_l = &[l, C_{l}, C_{l+1}, stride_l, k_l, 
		\\ &\text{FLOPs}[l],\text{FLOPs}_{1:l-1}, \text{FLOPs}_{l+1:L}, a_{l-1}]
	\end{aligned}
\end{equation}

\noindent where $k_l$ is the layer's kernel size.~$\text{FLOPs}_{1:l-1}$ denotes the number of previous layers' FLOPs given the actions that the model has done so far on them.~$\text{FLOPs}_{l+1:L}$ shows the next layers' FLOPs that are not pruned yet, and $a_{l-1}$ is the agent's action on the previous layer.

\noindent\textbf{Actions of the Agent:} Based on the state $S_l$ for the $l$-th layer, our agent determines its pruning rate $a_l$ such that $a_l \in [0, 1)$. Given the predicted pruning rate $a_l$, we remove $\lfloor a_l \times c_l \rfloor$ channels of the layer.~In addition, we calculate the minimum and maximum actual feasible pruning rates for the current layer based on $\text{FLOPs}_{1:l-1}$, $\text{FLOPs}[l]$, $\text{FLOPs}_{l+1:L}$, and the desired budget $\text{FLOPs}_{desire}$.~Then, we bound the predicted action $a_l$ to lie in the range $[a_{l, min}, a_{l, max}]$.~We refer to supplementary materials for more details.~In our experiments, we found that ranking the importance of filters using the norm criteria~\cite{li2016pruning} and pruning the ones with the lowest rank works well in our framework, but one may employ more sophisticated approaches~\cite{lin2020hrank} as well.

\noindent\textbf{Reward Function:} We set our reward function to be the pruned model's accuracy on a small held-out subset of the training dataset as a proxy for its final performance.~As our agent prunes one layer of the model at a time, it will be extremely time-consuming to calculate the proxy value after each action of the agent.~Thus, we take one pass of the agent on all layers of the model as one episodic trajectory for it. Then, we calculate the final pruned model's accuracy on the subset at the end of the trajectory and take it as the reward value for all state-action pairs seen during the trajectory.

\subsubsection{Modeling the Dynamic Nature of Rewards}\label{dynamic-rewards}
As we iteratively train the model's weights and the agent's policy, the weights of the convolution layers that the agent performs its actions on them are not static in our framework.~Thus, our reward function is dynamic in the course of training, which prohibits directly applying prominent RL methods~\cite{DBLP:journals/corr/LillicrapHPHETS15,haarnoja2018soft} that leverage episodic trajectories to train the agent's policy in our framework.~The reason is that the optimization procedure of these models is biased to only optimize the agent's policy \textit{w.r.t} the current episode's distribution and disregards the changing dynamics of the environment, resulting in a sub-optimal policy~\cite{khetarpal2022towards}.

We design a new mechanism to overcome this challenge by providing a representation of the dynamic environment to the agent. To do so, first,~we map the index of each epoch for training the model's weights to an embedding.~Then, we employ a recurrent GRU model~\cite{gru} that takes a sequence of the embeddings corresponding to the epochs that have been passed so far and outputs a representation of the current state of the model's weights.~Formally, if we train the model's weights for total $T$ epochs, we denote the epoch embeddings corresponding to epoch indexes $E = [e_1, e_2, \cdots, e_T]$ with $\Psi_{1:T} = [\psi_1,\psi_2,\cdots,\psi_T]$ ($\psi_1$ = Emb($e_1$), Emb is a learnable embedding layer).~We calculate the representation of the state of the model's weights at the epoch $e_k$ as follows:

\vspace{-8pt}
\begin{equation}\label{calc-z}
	z_k = f_{Env}(\Psi_{1:k}, h_0;\theta_{Env})
\end{equation}

\noindent$f_{Env}$ denotes the recurrent model.~$\Psi_{1:k}$ are the embeddings of epochs until the epoch $e_k$.~$h_0$ is the initial hidden state of the GRU that we set it to a zero vector, and $\theta_{Env}$ are the parameters of the GRU.~We show in section~\ref{rl-training} that we use the representations $z$ provided by the GRU for training our RL agent.

We propose to train the recurrent model using another model that we call it `decoder.' The decoder model takes 1) the state-action pairs $(S, a)$ and 2) the representation $z$ of the state of the environment when the agent observes $(S, a)$ and predicts the agent's reward $r$.~Our intuition is that the representation $z$ is informative of the state of the environment when the decoder can use it to accurately predict $r$.~We train both the recurrent model's weights and the ones for the decoder using the following objective:

\vspace{-10pt}
\begin{align}
	\min_{\theta_{Env}, \theta_D}\mathcal{L}_{recons} &= \mathbb{E}_{(s, a, r, e)\sim {B}}[(\hat{r} - r)^2]\label{recons-obj} \\ 
	\hat{r} &= f_D(S, a, z;\theta_D) \\
        z &= f_{Env}(\Psi, h_0; \theta_{Env})
\end{align}

\noindent In practice, we approximate the expectation in Eq.~\ref{recons-obj} using the agent's episodic observations during training.~$f_D$ is our decoder model, and $\theta_D$ represents its parameters.

\vspace{-8pt}
\subsubsection{RL Agent Training}~\label{rl-training}
We employ our recurrent model and the Soft Actor-Critic (SAC)~\cite{haarnoja2018soft} method to train our RL agent.~We augment the states $S$ with the representations of the environment's state $z$ and design our agent's policy function so that it predicts a distribution over actions conditioned on both of them:

\vspace{-10pt}
\begin{equation}\label{policy-define}
	a \sim \pi(\cdot | S, z~; \theta_A)
\end{equation}

\noindent Similarly, we deploy the representations $z$ when calculating the predicted Q-values by the critic networks in SAC.~We train them using the mean squared Bellman error objective:

\vspace{-15pt}
\begin{equation}\label{loss-critic}
	\small\mathcal{L}(\phi_i) = \mathbb{E}_{(s, a, r, s', d, e)\sim B}[(Q_{\phi_i}(s, a, z) - y(r, s', d, z))^2]
\end{equation}

\vspace{-15pt}
\begin{equation}
	\begin{aligned}
		\small&y(r, s', d, z) = r + \gamma(1 - d)[\min_{j=1,2}Q_{\phi_{targ, j}}(s', a', z)~- \\ 
		&\alpha\log(\pi(a'|s', z; \theta_A))];\\
  &a'\sim \pi(\cdot|s', z; \theta_A);~z = f_{Env}(\Psi, h_0)
	\end{aligned}
\end{equation}

\noindent$Q_{\phi_i}$ represents the critic models, and $Q_{\phi_{targ, i}}$ shows their target models obtained using Polyak averaging.~$B$ is a replay buffer containing previous episodic trajectories observed.~$(s,a)$ are state-action pairs from $B$.~$d$ indicates whether the state $s$ is a terminal state.~$s'$ represents the state that the model gets in after taking the action $a$ when being in the state $s$.~$a'$ is an action chosen using the most recent policy $\pi(\cdot;\theta_A)$ conditioned on the state $s'$ and environment representation $z$.~$\gamma$ is the discount factor for future rewards.~$\alpha$ determines the strength of the entropy regularization term, which is a hyperparameter.~Finally, we train the agent's policy using the following objective:

\vspace{-10pt}
\begin{equation}\label{loss-agent}
	\begin{aligned}
		\max_{\theta_A}\mathbb{E}_{(s, e)\sim B}[\min_{j=1,2}&Q_{\phi_j}(s, a) - \alpha\log\pi(a|s, z; \theta_A)] \\ 
		&a \sim \pi(a|s, z; \theta_A)
	\end{aligned}
\end{equation}

\begin{algorithm}
\SetAlgoLined
\SetNoFillComment
\small
\caption{Joint Training and Pruning}
\label{alg-1}

\KwIn{Training dataset $\mathcal{D}=\{(x_i, y_i)\}$; replay buffer $B$; CNN model $f_c(\cdot;~W)$ with $L$ layers; Agent   $\pi(\cdot;\theta_A)$; Two Critics $Q_{\phi_i}(\cdot)$ and their target models $Q_{\phi_{targ, i}}$ ($i \in \{1,2\}$); recurrent model $f_{Env}$ and epoch embeddings $\Psi$; regularization parameters $\alpha$, $\beta$; discount factor $\gamma$; number of iterations T; number of pruning episodes per iteration P; a subset $\mathcal{D}_s$ of $\mathcal{D}$ for calculating the agent's reward.}
 \For{$t := 1$ to $T$}{
    \tcc{Representation of Environment}
    1. Calculate $z_t$ using $f_{Env}$ and embeddings $\Psi$ in Eq.~(\ref{calc-z}).
    
    \tcc{RL Agent Exploration and Training}
    \For {$p := 1$ to $P$}{
        
        2.~Prune the $L$ layers of the CNN $f_c$ one at a time by calculating states $S_{l,p,t}$ and actions $a_{l,p,t}$ using $z_t$ and Eqs.~(\ref{state-define},\ref{policy-define}).
        
        3.~Calculate the reward $r_{p,k}$ using the final pruned model and $\mathcal{D}_s$. 
        
        4. Add the experiences ($S_{l,p,t}, a_{l,p,t}, S_{l+1,p,t}, e_t, r_{p,k}$) to the replay buffer $B$.
    }
    5.~Sample a batch of previous experiences $\mathcal{B}$ from $B$ and use them to calculate the loss value for the recurrent model, decoder, and epoch embeddings using Eq.~(\ref{recons-obj}).~Then, update their parameters using the Adam optimizer.
    
    6.~Use the samples in $\mathcal{B}$ to calculate the loss values of the critics $Q_{\phi_i}(\cdot)$ and the agent $\pi(\cdot;\theta_A)$ using Eqs.~(\ref{loss-critic},~\ref{loss-agent}).
    
    7.~Backpropagate the gradients of the calculated losses and update the parameters of the two critic models and the agent using the Adam optimizer.
    
    \tcc{Training the CNN's Weights}
    8.~Use the policy $\pi(\cdot)$ with the highest reward so far to determine the binary architecture vectors $[\textbf{v}_1$, $\textbf{v}_2$, $\cdots$, $\textbf{v}_L]$.
    
     9.~Calculate the loss $\mathcal{L}_w$ for the model's weights using Eqs.~(\ref{loss-align},~\ref{loss-weights}).~Backpropagate its gradients and update the model's parameters using SGD.
}
{\textbf{Return:}~Trained CNN model and agent.}
\end{algorithm}

\begin{table*}[t!]
\small
\caption{Comparison results on CIFAR-10 for pruning ResNet-56 and MobileNet-V2.}
\centering

\resizebox{0.67\textwidth}{!}{
\begin{tabular}{c|c|c|c|c|c}
\hline
Model & Method & Baseline Acc & Pruned Acc & $\Delta$-Acc & Pruned FLOPs \\ \hline
    %
    \multirow{10}{*}{\centering ResNet-56} &DCP-Adapt~\cite{zhuang2018discrimination}   &  $93.80\%$ & $93.81\%$ & $+0.01\%$ & $47.0\%$ \\ 
    &SCP~\cite{kang2020operation} &   $93.69\%$ & $93.23\%$ & $-0.46\%$ & $51.5\%$ \\ 
    &FPGM~\cite{he2019filter} &   $93.59\%$ & $92.93\%$ & $-0.66\%$ & $52.6\%$ \\
    &SFP~\cite{he2018soft} &   $93.59\%$ & $92.26\%$ & $-1.33\%$ & $52.6\%$ \\
    &AMC~\cite{he2018amc} &   $92.80\%$ & $91.90\%$ & $-0.90\%$ & $50.0\%$ \\
    &FPC~\cite{he2020learning} &   $93.59\%$ & $93.24\%$ & $-0.25\%$ & $52.9\%$
    \\
    &HRank~\cite{lin2020hrank} &  $93.26\%$ & $92.17\%$ & $-0.09\%$ & $50.0\%$ \\
    &DTP~\cite{li2023DifferentiableTransportationPruning} &  $93.36\%$ & $93.46\%$ & $+0.10\%$ & $50.0\%$ \\
    & RLAL (ours) & $93.41\%$ & $\mathbf{93.86\%}$ & $\bm{+}$ $\mathbf{0.45\%}$ & $50.0\%$ \\
    \hline
    \multirow{6}{*}{MobileNet-V2}&Uniform~\cite{zhuang2018discrimination}&$94.47\%$& $94.17\%$&$-0.30\%$&$26.0\%$ \\
    &SCOP~\cite{tang2020scop}&$94.48\%$& $94.24\%$ &$-0.24\%$&$26.0\%$\\
    &MDP~\cite{guo2020multi}&$95.02\%$& $95.14\%$ &$+0.12\%$&$28.7\%$\\
    &DCP~\cite{zhuang2018discrimination}&$94.47\%$& $94.69\%$ &$+0.22\%$&$40.3\%$ \\  
    &DDNP~\cite{gao2022disentangled} &$94.58\%$&$94.81\%$ &$\bm{+}$ $0.23\%$&$43.0\%$ \\
    &RLAL (ours) &$94.48\%$&$\mathbf{94.85\%}$ &$\bm{+}$ $\mathbf{0.37\%}$&$\mathbf{49.4\%}$ \\
      
    \hline
\end{tabular}}
\label{tab:cifar10Cmp}
\end{table*}

\vspace{-10pt}
\subsubsection{Soft Regularization of the Model's Weights}~\label{soft-reg}
As mentioned in the section~\ref{iterative-training}, we iteratively train and prune the model's weights in our framework.~One approach to do so can be actually pruning the model's architecture by removing the redundant channels selected by the agent and only training the remaining ones in the weight training phase.~However, doing so can make the training procedure unstable because it can significantly drop the model's accuracy.~Accordingly, we propose an alternative approach to softly regularize the model's weights to align with the selected sub-network by the agent.~Given binary architecture vectors [$\textbf{v}_1$, $\textbf{v}_2$, $\cdots$, $\textbf{v}_L$] denoting the channels selected by the current \underline{\textbf{best}} agent for each layer, we use the following regularization term to train the model's weights:

\begin{equation}\label{loss-align}
	\mathcal{L}_{align} = \sum_{l=1}^{L}||(1 - \textbf{v}_l) \odot \mathcal{W}_l||_2
\end{equation}

\noindent Here, $\odot$ means element-wise product and the proposed objective applies the Group Lasso regularization on the channels removed by the agent.~Finally we combine the standard Cross Entropy Loss ($\mathcal{L}_{class}$) with proposed $\mathcal{L}_{align}$ to train the model's weights:

\begin{equation}\label{loss-weights}
	\mathcal{L}_w = \mathcal{L}_{class} + \beta\mathcal{L}_{align}
\end{equation}

\noindent In practice, we apply $\mathcal{L}_{align}$ using the version of the policy with the highest reward until the current training iteration to make the training more stable.~We summarize our training algorithm for training the model's weights and optimizing the agent's policy in Alg.~\ref{alg-1}.

\section{Experiments}
\vspace{-5pt}
We conduct experiments on ImageNet~\cite{deng2009imagenet} and CIFAR-10~\cite{krizhevsky2009learning} to analyze the performance of our method.~For all experiments, we use fully connected models with two hidden layers of size $300$ for the architecture of the actor, two critics, and two target models of the critic models.~We train the agent and critic models using the Adam optimizer~\cite{DBLP:journals/corr/KingmaB14} with learning rate of $1\mathrm{e}{-4}$ and $1\mathrm{e}{-3}$ respectively.~We use exponential decay rates of $(\beta_1,\beta_2)=(0.9,0.999)$ for all of them.~For our recurrent model, we use a GRU~\cite{gru} model with the input size of $128$.~We also take embeddings of size $128$ for all epochs~(Sec.~\ref{dynamic-rewards}).~We employ a fully connected model we two hidden layers of size $300$ as our decoder model.~We train the GRU model, epoch embeddings, and the decoder model using the Adam optimizer with learning rate of $1\mathrm{e}{-3}$ and decay parameters of $(\beta_1, \beta_2) = (0.9, 0.999)$.~We set the entropy regularization coefficient $\alpha$ to 0.1 and $\beta$ for soft regularization to $1\mathrm{e}{-4}$ for all models.~Finally, we choose the number of episodic observations per epoch for our agent to be $P=10$ (see Alg.~\ref{alg-1}).~In all experiments, as we jointly train and prune our model by using \textbf{\underline R}einforcement \textbf{\underline L}earning and softly \textbf{\underline{AL}}igning the weights of the model with the selected sub-networks, we call our method~\textbf{RLAL}.~We refer to supplementary materials for more details of our experiments.

\begin{table*}[t!]
\small
\centering
\caption{\small Comparison results on ImageNet for pruning ResNet-18/34 and MobileNet-V2.}
\resizebox{0.82\textwidth}{!}{
\begin{tabular}{c|c|c|c|c|c|c}
    \hline
    Model & Method  & Baseline Top-1 Acc & Baseline Top-5 Acc &  $\Delta$-Acc Top-1 & $\Delta$-Acc Top-5 & Pruned FLOPs \\ \hline
    \multirow{9}{*}{ResNet-18} &
    MIL~\cite{dong2017more}&$70.28\%$ &$89.63\%$&$-3.18\%$&$-1.85\%$&$ 41.8\%$\\
    &SFP~\cite{he2018soft}&$70.28\%$ &$89.63\%$&$-3.18\%$&$-1.85\%$&$ 41.8\%$\\
    &FPGM~\cite{he2019filter}&$70.28\%$ &$89.63\%$&$-1.87\%$&$-1.15\%$&$ 41.8\%$\\
    &PFP~\cite{liebenweinprovable}&$69.74\%$ &$89.07\%$&$-2.36\%$&$-1.16\%$&$29.3\%$\\
    &SCOP~\cite{tang2020scop}&$69.76\%$ &$89.08\%$&$-1.14\%$&$-0.93\%$&$ 45.0\%$\\
    &GNNRL~\cite{yu2022topology}&$69.76\%$ & - &$-1.10\%$& - & $51.0\%$\\
    &GP~\cite{herrmann2020channel} & $70.28\%$ & $89.63\%$ & $-1.40\%$ & $-0.97\%$ & ${43.9\%}$ \\
    &PGMPF~\cite{cai2022PGMB} & $70.23\%$ & $89.51\%$ & $-3.56\%$ & $-2.15\%$ & ${53.5\%}$ \\
    &FTWT~\cite{elkerdawy2022FTWT} & $69.76\%$ & - & $-2.27\%$ & - & ${51.5\%}$ \\
    &EEMC~\cite{zhang2021exploration} & $70.28\%$ & $89.63\%$ & $-2.01\%$ & $-1.19\%$ & ${46.6\%}$ \\
    & RLAL (Ours) &   $69.80\%$ & $89.10\%$ & $\mathbf{-0.80\%}$ & $\mathbf{-0.42\%}$ & $\mathbf{50.0\%}$ \\
    \hline

    \multirow{11}{*}{ResNet-34}  &
     SFP~\cite{he2018soft} &  $73.92\%$ & $91.62\%$ & $-2.09\%$ & $-1.29\%$ & $56.0\%$ \\
    &FPGM~\cite{he2019filter}&$73.92\%$& $91.62\%$& $-1.29\%$& $-0.54\%$& $41.1\%$\\
    &Taylor~\cite{molchanov2019importance}&$73.31\%$ &-&$-0.48\%$&-&$24.2\%$\\
    & SCOP~\cite{tang2020scop} &  $73.31\%$ & $91.42\%$ & $-0.69\%$ & $-0.44\%$ & $44.8\%$ \\
     &GP~\cite{herrmann2020channel}&$73.92\%$& $91.62\%$& $-1.14\%$& $- 0.69\%$& $51.1\%$\\
    & DMC~\cite{gao2020discrete} &  $73.30\%$ & $91.42\%$ & $-0.73\%$ & $-0.31\%$ & $43.4\%$ \\
    & PGMPF~\cite{cai2022PGMB} &  $73.27\%$ & $91.43\%$ & $-1.68\%$ & $-0.98\%$ & $52.7\%$ \\
    & FTWT~\cite{elkerdawy2022FTWT} &  $73.30\%$ & - & $-1.59\%$ & - & $52.2\%$ \\
    & GFS~\cite{ye2020good} &  $73.31\%$ & - & $-0.40\%$ & - & $43.8\%$ \\
    & ISP~\cite{ganjdanesh2022interpretations} &  $73.31\%$ & $91.42\%$ & $-0.45\%$ & $-0.40\%$ & $44.0\%$ \\
    & RLAL (Ours) &  $73.45\%$ & $91.48\%$ & $\bm{-}$ $\mathbf{0.14\%}$ & $\bm{-}$ $\mathbf{0.23\%}$ & $\mathbf{50.0\%}$ \\
    \hline
    
    \multirow{6}{*}{MobileNet-V2} & Uniform~\cite{sandler2018mobilenetv2} &  $71.80\%$ & $91.00\%$ & $-2.00\%$ & $-1.40\%$ & $30.0\%$ \\
     &AMC~\cite{he2018amc} &  $71.80\%$ & - & $-1.00\%$ & - & $30.0\%$ \\
     &Random~\cite{li2022revisiting} & $71.88\%$ & - & $-0.98\%$ & - & $28.9\%$ \\
     &CC~\cite{li2021towards} &  $71.88\%$ & - & $-0.97\%$ & - & $28.3\%$ \\
    &MetaPruning~\cite{liu2019metapruning}   & $72.00\%$ & - & $-0.80\%$ & - & $\mathbf{30.7\%}$ \\
    & RLAL (ours)  & $71.82\%$ & $90.26\%$ & $\bm{-}$ $\mathbf{0.50\%}$ & $\bm{-}$ $\mathbf{0.33\%}$ & ${29.4\%}$ \\\hline
    \end{tabular}}

\label{tab:imgnetCmp}
\vspace{-5pt}
\end{table*}

\subsection{CIFAR-10 Results}
Tab.~\ref{tab:cifar10Cmp} summarizes comparison results on the CIFAR-10 dataset.~As can be seen, for ResNet-56, RLAL can achieve the best accuracy~\textit{vs.}~computational efficiency trade-off compared to the baseline methods.~One the one hand, it is able to prune FLOPs with a rate comparable to ($<3\%$ lower) FPC~\cite{he2020learning} while achieving $+0.70$ higher $\Delta$-Acc.~On the other hand, only RLAL, DTP~\cite{li2023DifferentiableTransportationPruning}, and DCP-Adapt are able to outperform their baseline methods.~RLAL can both prune $3\%$ more FLOPs and accomplish $0.44\%$ better $\Delta$-Acc than DCP-Adapt.~It also has $0.44\%$ higher $\Delta$-Acc than DTP with the same FLOPs reduction ratio.~Finally, with the same FLOPs pruning rate, RLAL significantly outperforms AMC~\cite{he2018amc} with $1.35\%$ higher $\Delta$-Acc.~For MobileNet-V2, RLAL can attain the highest $\Delta$-Acc while having the largest pruning rate at the same time.~It remarkably prunes $6.4\%$ more FLOPs than DDNP~\cite{gao2022disentangled} while obtaining $0.14\%$ higher $\Delta$-Acc.~In summary, these results demonstrate the effectiveness of our method for finding efficient yet accurate models.

\subsection{ImageNet Results}
We present the experimental results on ImageNet in Tab.~\ref{tab:imgnetCmp}.~For ResNet-18, RLAL shows the best $\Delta$-Acc Top-$1/5$ while showing a competitive pruning rate.~It has a similar pruning rate (only $1\%$ lower) to GNNRL~\cite{yu2022topology} and achieves $0.30\%$ higher $\Delta$-Acc Top-$1$.~For pruning ResNet-34, RLAL is able to find a proper balance between accuracy and efficiency of the model.~For instance, with a similar computation budget to GP~\cite{herrmann2020channel} (only $1.1\%$ FLOPs difference), RLAL's pruned model has $1\%/0.46\%$ higher $\Delta$-Acc Top-$1/5$.~Moreover, RLAL shows better final accuracies than ISP~\cite{ganjdanesh2022interpretations} while significantly pruning more FLOPs with a $6\%$ margin.~Pruning MobileNet-V2 is more challenging compared to ResNets because MobileNets~\cite{howard2017mobilenets,sandler2018mobilenetv2} are primarily designed for efficient inference.~Accordingly, improvements in metrics are more difficult to secure than ResNet cases.~We can observe that all methods have close FLOPs pruning rates in a relatively small range from $28.3\%$ to $30.7\%$.~RLAL can achieve $0.5\%$ higher $\Delta$-Acc Top-$1$ while pruning only $0.6\%$ lower FLOPs compared to AMC, and it has the best $\Delta$-Acc Top-$1$ among baselines.~These results illustrate the capability of our method to effectively prune both large and small size models.~Further, we highlight the advantages of our method compared to baseline RL-based pruning methods, GNNRL~\cite{yu2022topology} and AMC~\cite{he2018amc}, as it can obtain more accurate pruned models while not requiring a pretrained model for pruning.

\begin{figure*}[tbh!]
\vspace{-10pt}
    \centering
        \subfloat[Pruning Rate: $35\%$]{
		\begin{minipage}[b]{.27\linewidth}
			\centering
			\includegraphics[width=\textwidth]{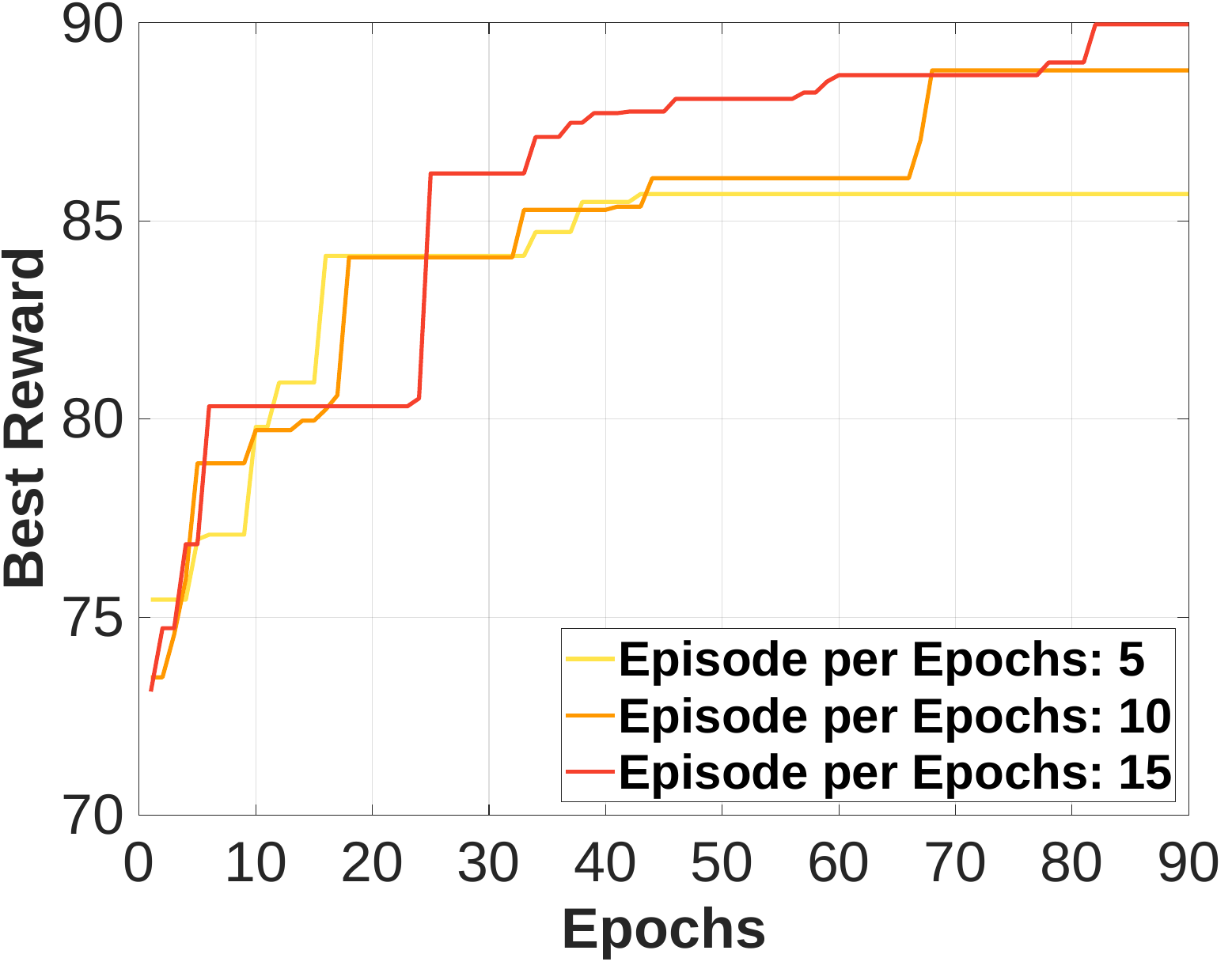}
	\end{minipage}}
        \hspace{0.02\linewidth}
	\subfloat[Pruning Rate: $50\%$]{
		\begin{minipage}[b]{.27\linewidth}
			\centering
			\includegraphics[width=\textwidth]{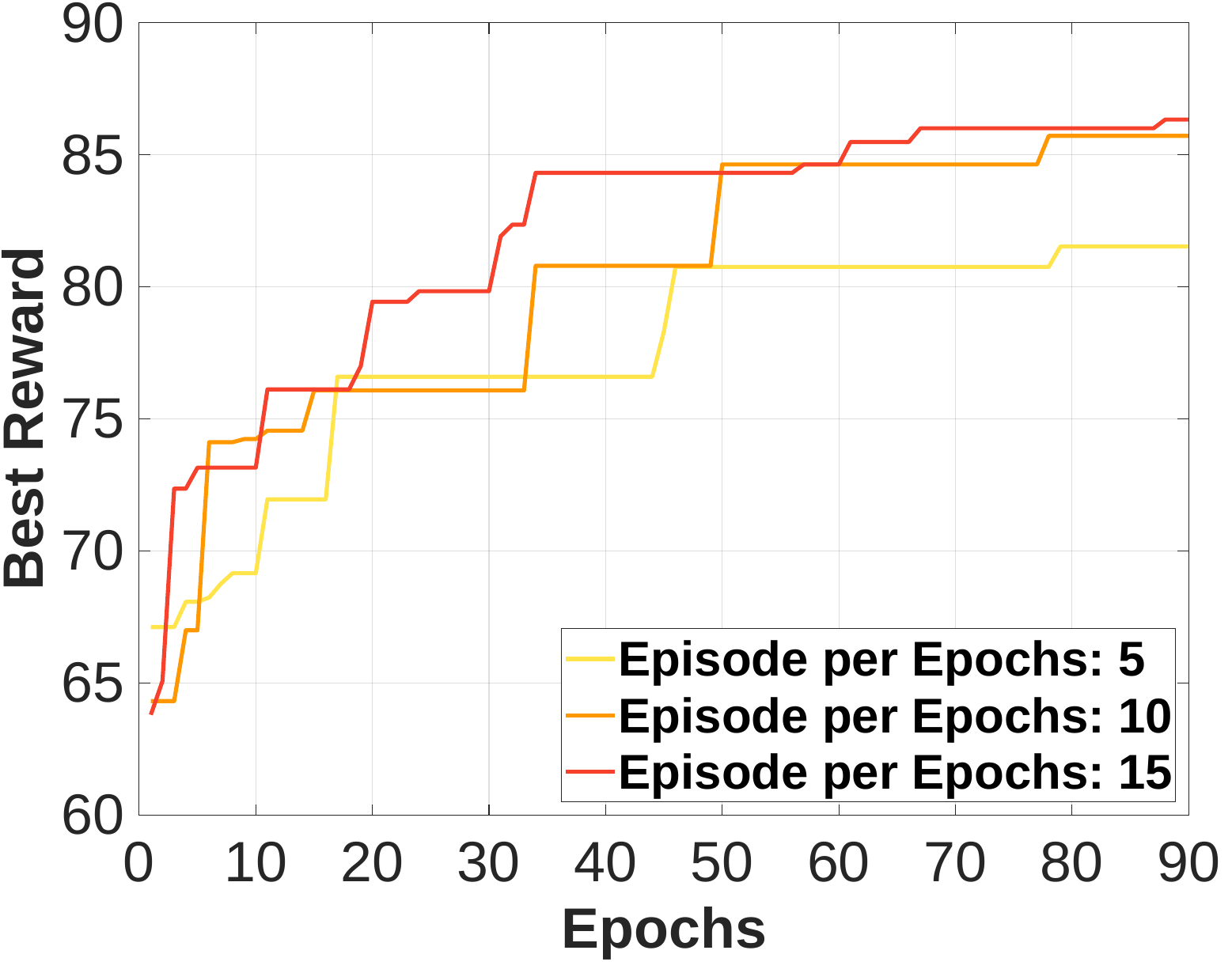}
	\end{minipage}}
        \hspace{0.02\linewidth}
        \subfloat[Pruning Rate: $65\%$]{
	\begin{minipage}[b]{.27\linewidth}
			\centering
			\includegraphics[width=\textwidth]{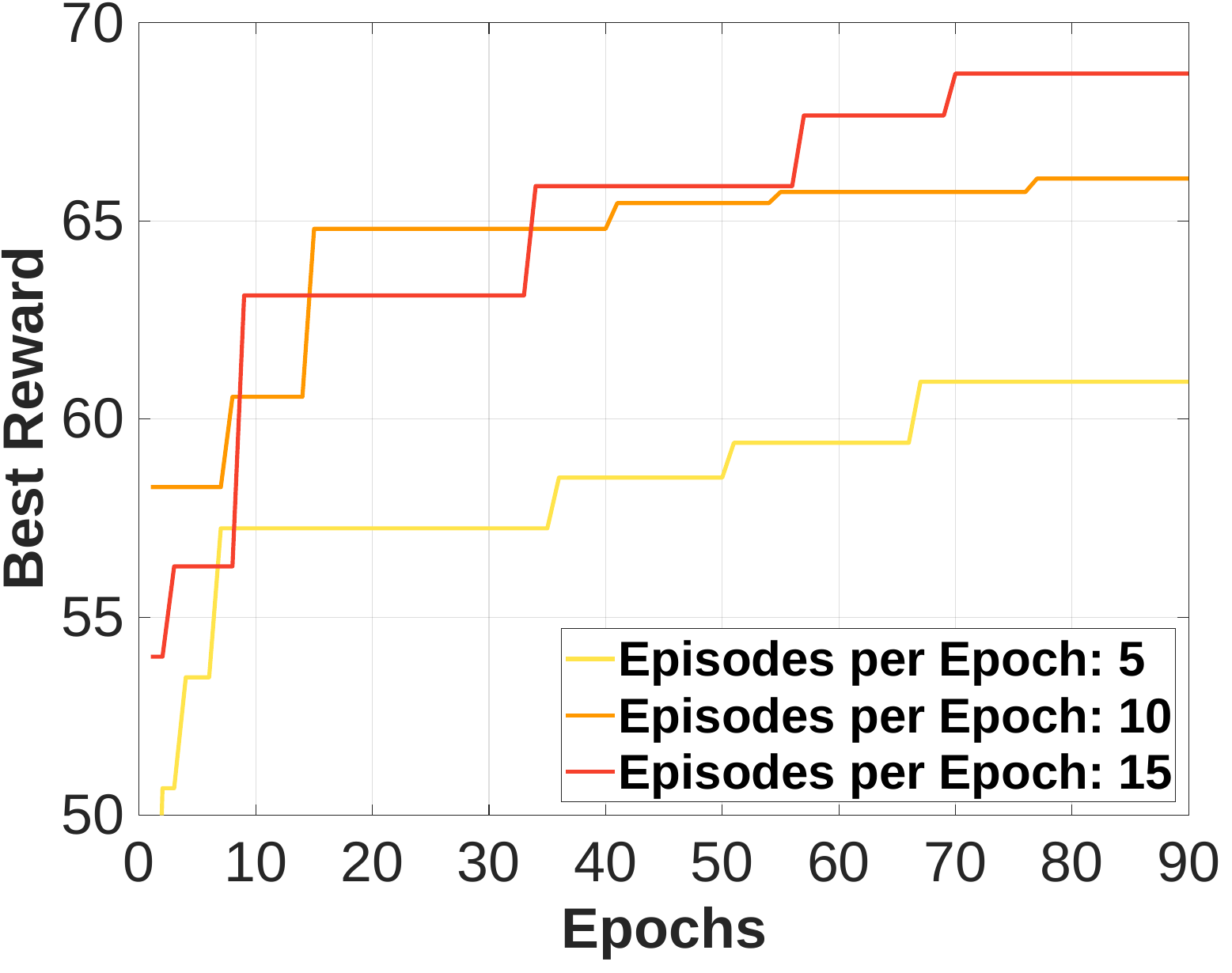}
	\end{minipage}}
	\\
	\subfloat[Pruning Rate: $35\%$]{
		\begin{minipage}[b]{.27\linewidth}
			\centering
			\includegraphics[width=\textwidth]{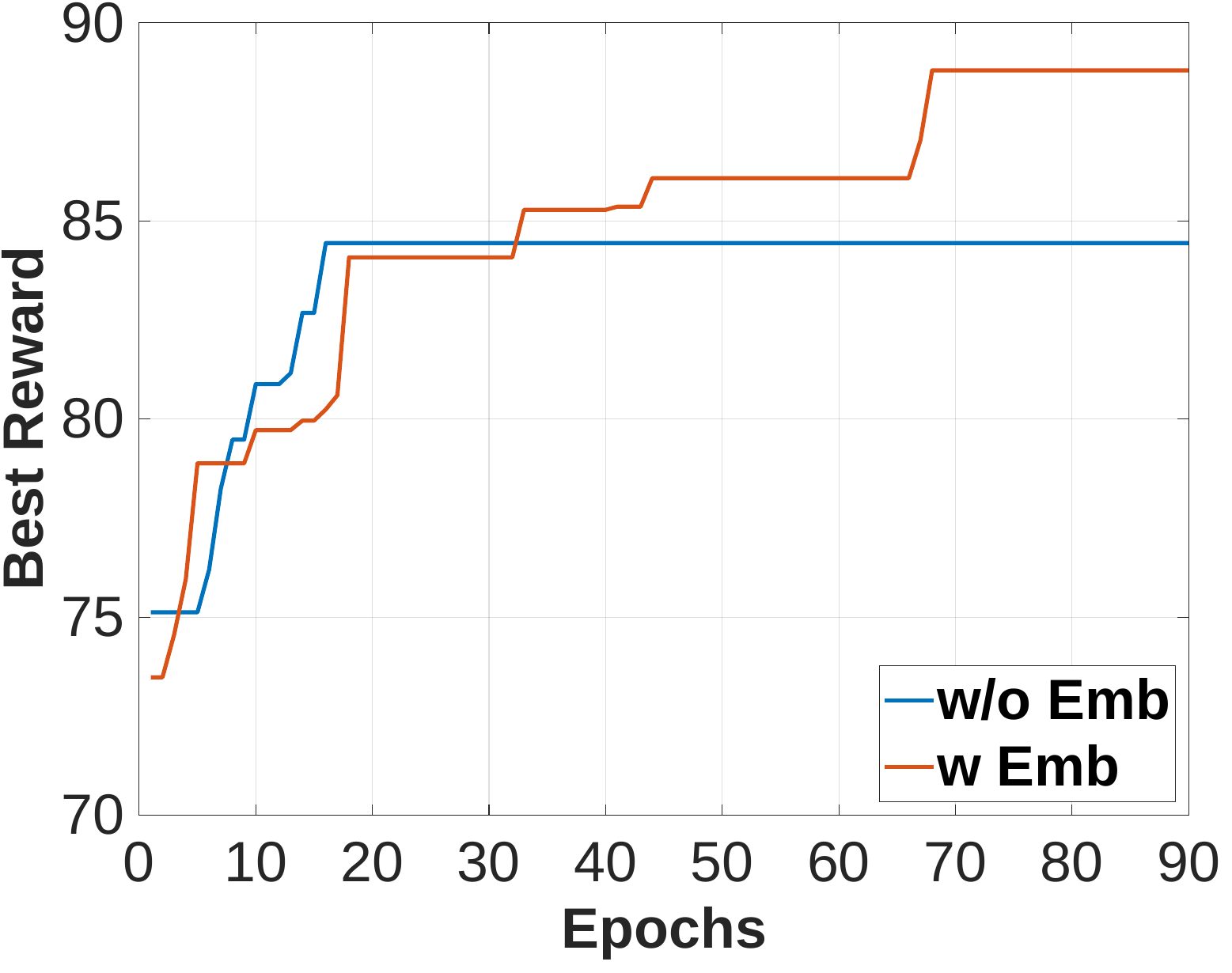}
	\end{minipage}}
        \hspace{0.02\linewidth}
	\subfloat[Pruning Rate: $50\%$]{
		\begin{minipage}[b]{.27\linewidth}
			\centering
			\includegraphics[width=\textwidth]{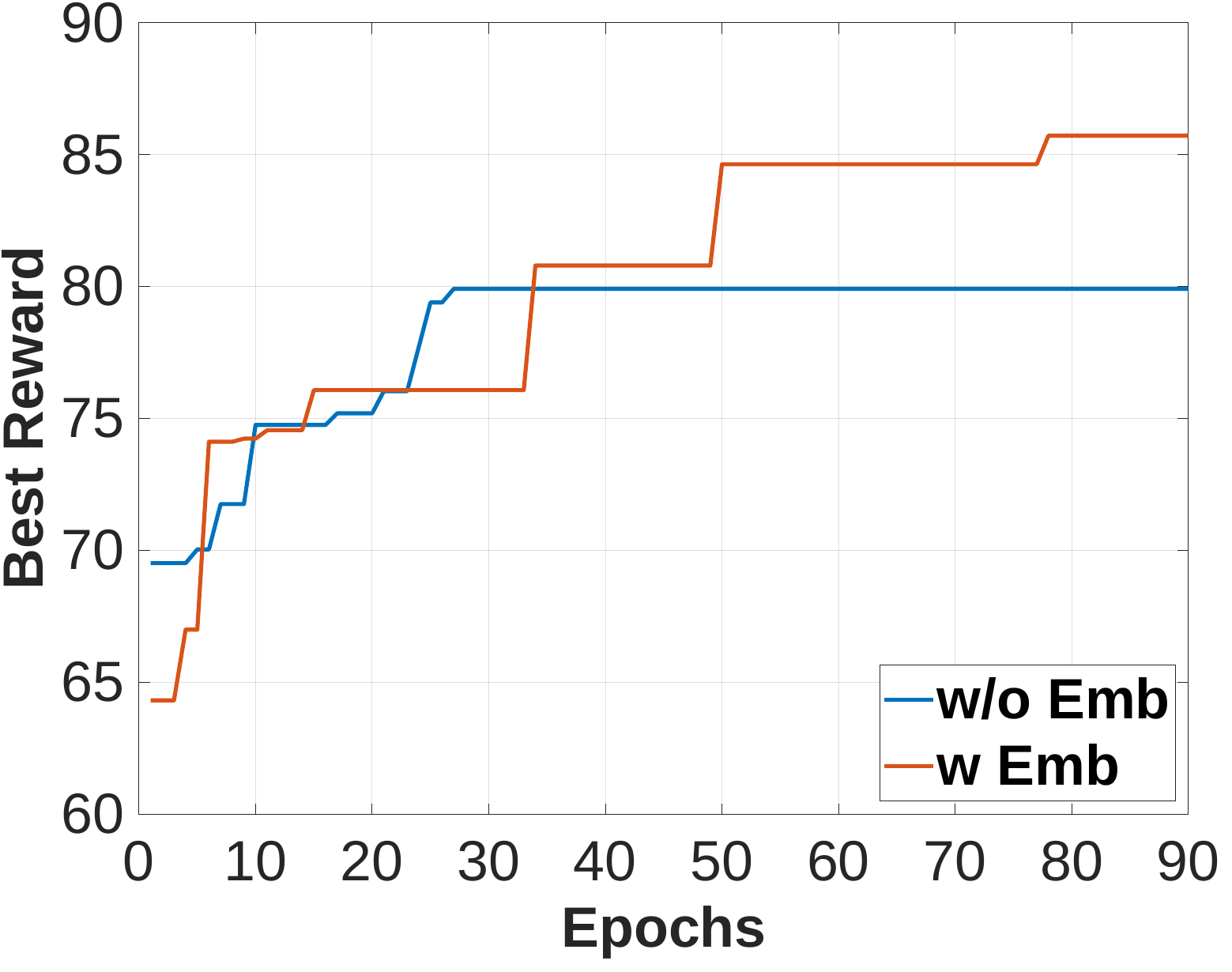}
	\end{minipage}}
        \hspace{0.02\linewidth}
        \subfloat[Pruning Rate: $65\%$]{
	\begin{minipage}[b]{.27\linewidth}
			\centering
			\includegraphics[width=\textwidth]{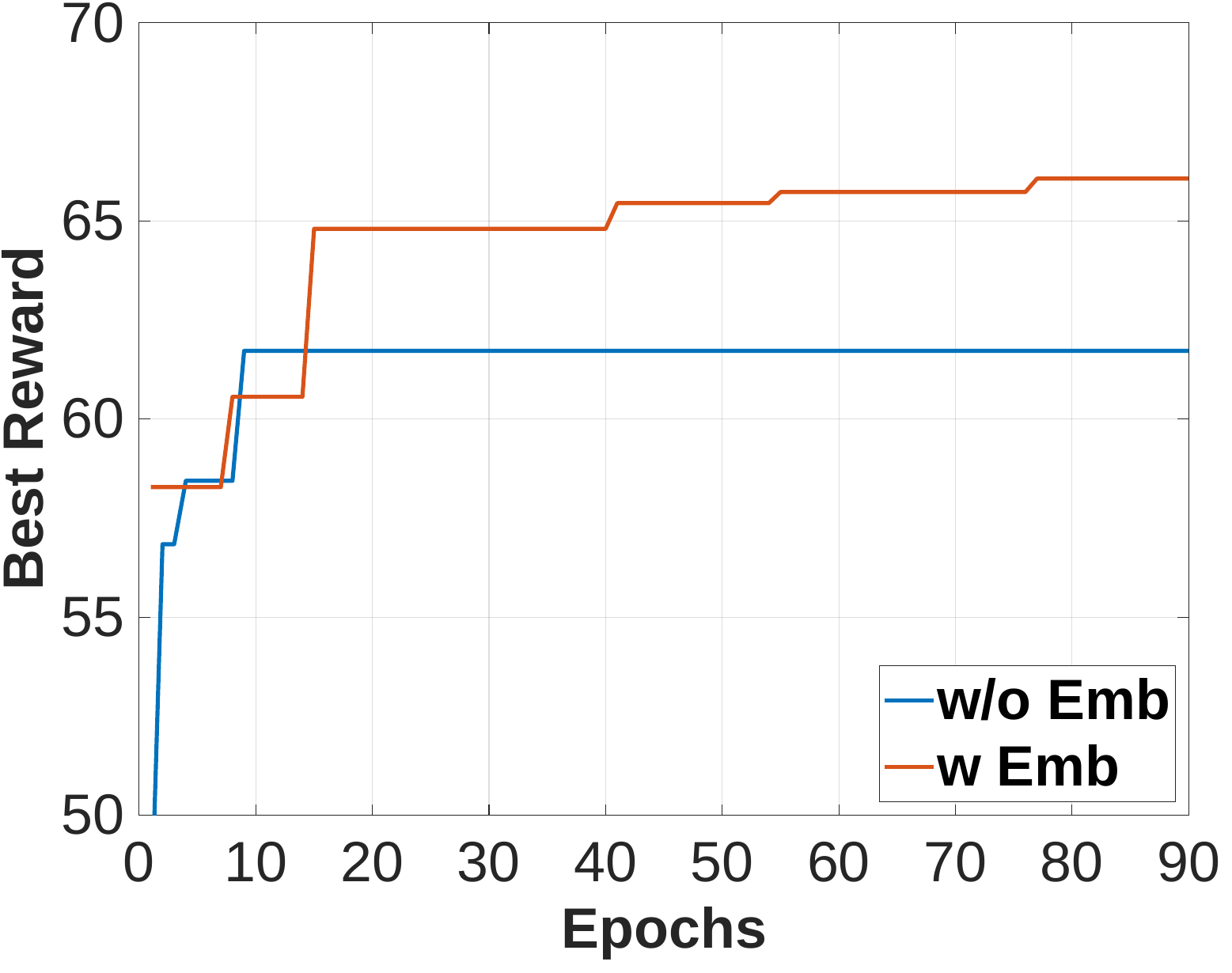}
	\end{minipage}} \\

        \subfloat[Pruning Rate: $35\%$]{
		\begin{minipage}[b]{.27\linewidth}
			\centering
			\includegraphics[width=\textwidth]{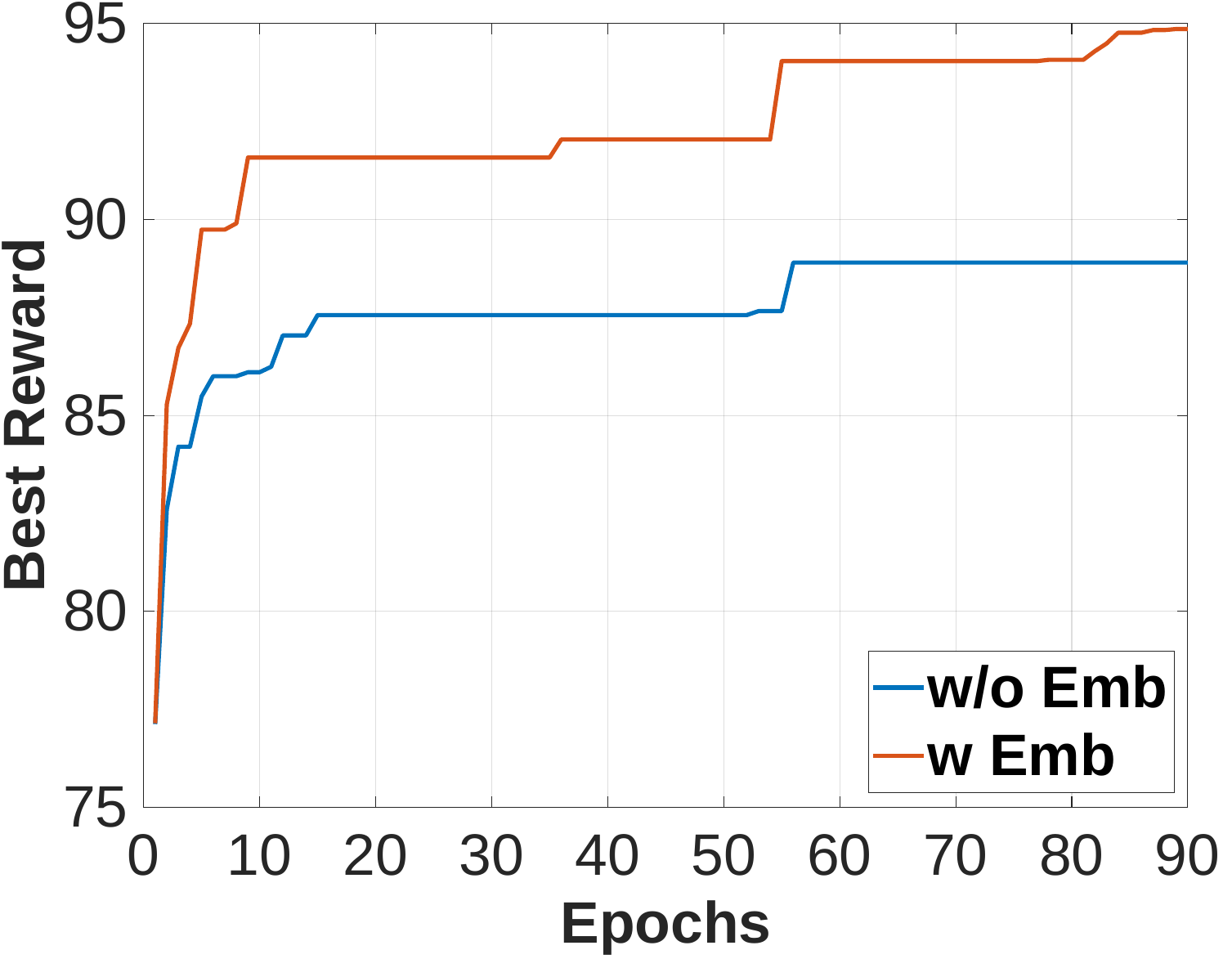}
	\end{minipage}}
        \hspace{0.02\linewidth}
	\subfloat[Pruning Rate: $50\%$]{
		\begin{minipage}[b]{.27\linewidth}
			\centering
			\includegraphics[width=\textwidth]{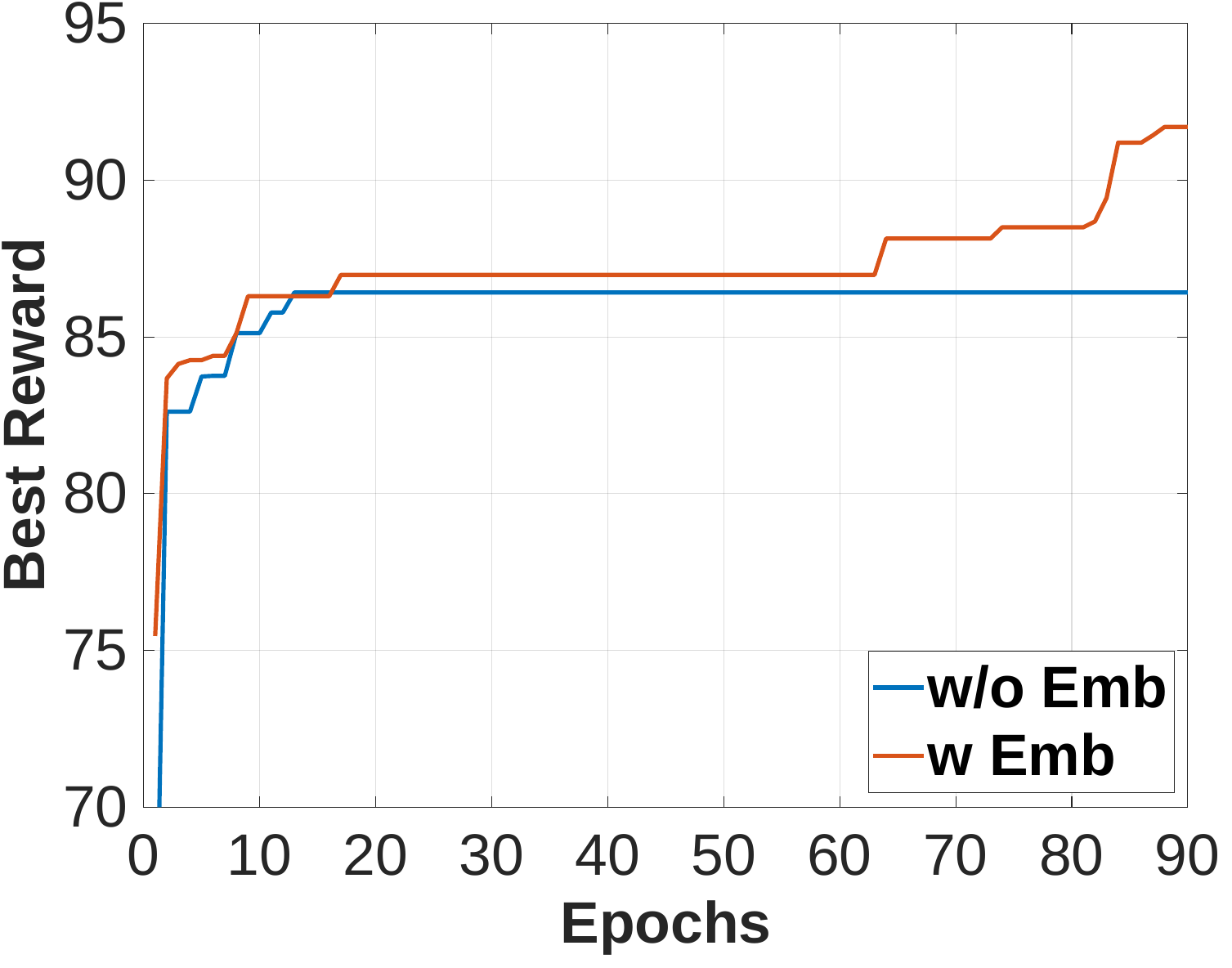}
	\end{minipage}}
 \hspace{0.02\linewidth}
        \subfloat[Pruning Rate: $65\%$]{
	\begin{minipage}[b]{.27\linewidth}
			\centering
			\includegraphics[width=\textwidth]{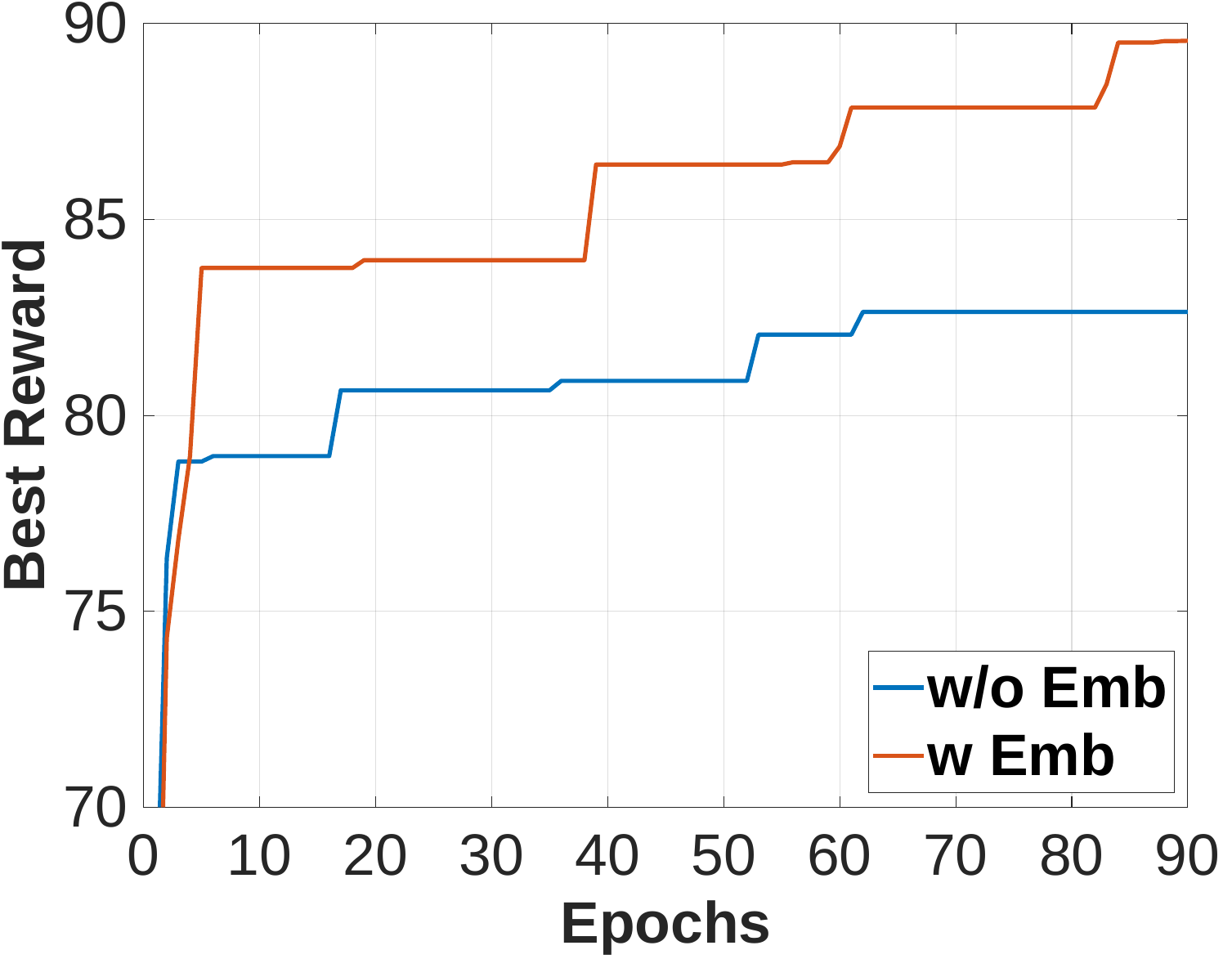}
	\end{minipage}}
    \caption{\small Results of ablation experiments on CIFAR-10.~\textbf{(a-c)}~Best reward of our agent when using a different number of episodes per epoch for three pruning rates when pruning ResNet-56.~\textbf{(d-f)}~Best reward with/without using our mechanism to provide representations of the environment to our agent during training for three pruning rates for ResNet-56.~\textbf{(g-i)}~Same results of~\textbf{(d-f)} for MobileNet-V2.}
    \label{fig:best_reward_stat}
    \vspace{-10pt}
\end{figure*}

\subsection{Ablation Studies}

We conduct ablation experiments to explore our method's behavior by studying 1) the effect of changing the number of episodic observations of the agent in each epoch and 2) the advantage of using our soft regularization, epoch embeddings, and the recurrent environment model in our framework.~We refer to supplementary materials for details of experimental settings.

\noindent\textbf{Changing the Number of Episodes:}~We experiment using a ResNet-56~\cite{he2016deep} model on CIFAR-10 with three different pruning rates in $\{35\%, 50\%, 65\%\}$, and we set the number of episodic observations for our agent in each epoch from $\{5, 10, 15\}$.~For each pruning ratio, we visualize the best reward that the agent achieves during the training~\textit{vs.}~the epoch numbers.~The results are shown in Fig.~\ref{fig:best_reward_stat}~(a-c).~We can observe a common trend in all cases that increasing the number of episodes results in a higher final reward, especially, the higher number of episodes benefits more when the desired compression ratio is larger at the cost of longer training time.~However, if the number of episodes is large enough, our method can attain a decent final reward value in a reasonable time.

\begin{table}[]
\small
\centering
\caption{Ablation Results of our method for pruning ResNet-56 on the CIFAR-10 dataset.~EE represents the \textbf{E}poch \textbf{E}mbeddings.~SR represents the \textbf{S}oft \textbf{R}egularization in Eq.~\ref{loss-align}.}
\resizebox{0.95\linewidth}{!}{
\begin{tabular}{c|c|c|c|c}
\hline
Setting & \begin{tabular}[c]{@{}c@{}}Baseline\\ Acc\end{tabular} & \begin{tabular}[c]{@{}c@{}}Pruned\\ Acc\end{tabular} & \begin{tabular}[c]{@{}c@{}}$\Delta$-Acc\end{tabular} & \begin{tabular}[c]{@{}c@{}}Pruned\\ FLOPs\end{tabular} \\ \hline
\begin{tabular}[c]{@{}c@{}}w/o EE\end{tabular} & $93.47\%$ & $93.44\%$ & $-0.03\%$ & \multirow{3}{*}{50\%} \\ \cline{1-4}
\begin{tabular}[c]{@{}c@{}}w/o EE + w/o SR \end{tabular} & $93.33\%$ & $93.12\%$ & $-0.21\%$ & \\ \cline{1-4}
Ours & $93.41\%$ & $93.86\%$ & $+0.45\%$ & \\ \hline \end{tabular}}
\label{ablation-results}
\end{table}

\noindent\textbf{Benefit of the Recurrent Environment Model:}~In our second experiment, we prune and finetune ResNet-56 and MobileNet-V2~\cite{sandler2018mobilenetv2} with three pruning rates in $\{35\%, 50\%, 65\%\}$ while using/dropping our mechanism to provide a representation of the environment to the agent using the epoch embeddings and our recurrent model.~We visualize the best reward of the agent in the course of training.~The results for ResNet-56 and MobileNet-V2 are shown in Fig.~\ref{fig:best_reward_stat}~(d-f) and Fig.~\ref{fig:best_reward_stat}~(g-i), respectively.~The cases using/not using our mechanism are shown with `w Emb'/`w/o Emb.'~The results clearly demonstrate the benefit of our design that provides a representation of the environment to the agent.~We can find that `w/o Emb' cases commonly reach to a relatively high reward but cannot properly deal with the dynamic reward function for their agent to further improve their policy.~In contrast, our method can consistently enhance its policy to reach higher reward values during training.

In our third experiment, we prune ResNet-56 on CIFAR-10 with two settings: 1) not using the recurrent model and epoch embeddings to provide representations of the environment to the agent 2) neither using the recurrent model and epoch embeddings nor the soft regularization.~We present the results in Tab.~\ref{ablation-results}.~One can notice that removing each component of our method degrades its performance, especially not using the recurrent model and epoch embeddings severely degrades our method's accuracy, which is inline with the results presented in Fig.~\ref{fig:best_reward_stat} and discussed above.~In summary, our ablation studies illustrate the effectiveness of our design choices in our method for jointly training and pruning a CNN model.

\section{Conclusion}
We proposed a method for structural pruning of a CNN model that jointly trains its weights and prunes its channels using an RL agent.~Our method iteratively updates the model's weights and allows the agent to observe several episodic pruning trajectories that it performs on the model to update its policy.~Our agent's actions determine the pruning ratios for the layers of the model, and we set the resulting model's accuracy to be the agent's reward.~Such a design brings about a dynamic reward function for the agent.~Thus, we developed a mechanism to model the complex dynamics of the reward function and yield a representation of it to the agent. To do so, we mapped the index of each epoch of the training to an embedding.~Then, we employed a recurrent model that takes the embeddings and provides a representation of the evolving environment's state to the agent.~We train the recurrent model and embeddings by utilizing a decoder model that predicts the agent's rewards given observed states, actions, and environment representations predicted by the recurrent model.~Finally, we regularized the model's weights to align with the sub-network selected by the agent's policy with the highest reward during training.~Our designs enable the agent to effectively leverage the environment representations along with its episodic observations to learn a proper policy for pruning the model while interacting in our non-stationary pruning environment.~Our experiments on ImageNet and CIFAR-10 demonstrate that our method can achieve competitive results with prior methods, especially the ones that use RL for pruning, while not requiring a pretrained model before pruning like them.

{
    \small
    \bibliographystyle{ieeenat_fullname}
    \bibliography{main}
}

\appendix

\newpage

\section{Bounding our Agent's Actions}
As mentioned in Section 3.3 of our paper, we calculate the minimum ($a_{l,min}$) and maximum ($a_{l,max}$) feasible pruning rates for the $l$-th layer before pruning it to ensure that reaching the desired FLOPs budget,~$\text{FLOPs}_{desire}$, is still possible after doing so.~However, before formally introducing our scheme for calculating $a_{l,min}, a_{l,max}$, we present how we implement our pruning actions in practice.

\subsection{Implementation of our Agent's Actions}
We describe our implementation for our agent's actions for each architecture.~For all models, we take each block of a CNN model as one `layer' in our framework.

\noindent\textbf{ResNets}: for our experiments on ResNet~\cite{he2016deep} models~(ResNet-56 on CIFAR-10~\cite{krizhevsky2009learning} and ResNet-18/34 on ImageNet~\cite{deng2009imagenet}), we take each residual block as one layer. It contains a structure as \texttt{Conv1-BN-ReLU-Conv2-BN} where \texttt{Conv1} and \texttt{Conv2} are the convolution layers, \texttt{BN} represents Batch Normalization~\cite{ioffe2015batch}, and \texttt{ReLU} is the ReLU activation function.~For each block, given the predicted action $a_l$ for pruning it, we remove $\lfloor a_l \times c \rfloor$ output channels of the \texttt{Conv1} layer and the same number of input channels of the \texttt{Conv2} layer where $c$ is the number of output/input channels of \texttt{Conv1}/\texttt{Conv2}.

\noindent\textbf{MobileNet-V2}:~for~experiments~using~MobileNet-V2, we take each inverted residual block~\cite{sandler2018mobilenetv2} as one layer for pruning.~Each block has the structure with \texttt{Conv1-BN-ReLU6-DW\_Conv-BN-ReLU6-Conv2-BN} form where \texttt{DW\_Conv} is a depth-wise convolution layer.~Given the predicted action $a_l$, we remove $\lfloor a_l \times c \rfloor$ output channels of \texttt{Conv1} and the same amount of channels of \texttt{DW\_Conv} and input channels of \texttt{Conv2}.

In summary, our pruning scheme changes the inner number of channels in each block of a CNN and preserves its number of input and output channels.

\subsection{Calculating Action Bounds}
We calculate $a_{l,min}, a_{l,max}$ for the $l$-th layer based on the total model's FLOPs that we denote with $\text{FLOPs}_T$ the number of FLOPs for the previous pruned layers $\text{FLOPs}_{1:l-1}$; the number of FLOPs for the next remaining layers $\text{FLOPs}_{l+1:L}$; $\text{FLOPs}[l]$; and $\text{FLOPs}_{desire}$.~The formulations are as follows:

\begin{equation}
    a_{l,min} = 1 - \frac{\text{FLOPs}_{desire} - \text{FLOPs}_{1:l-1}}{\text{FLOPs}[l]}
\end{equation}

\begin{equation}
    a_{l,max} = 1 - \frac{\text{FLOPs}_{desire} - \text{FLOPs}_{1:l-1} - \text{FLOPs}_{l+1:L}}{\text{FLOPs}[l]}
\end{equation}

In these equations, $a_{l,max}$ prevents very high pruning rates that even if all the next layers are kept intact, reaching $\text{FLOPs}_{desire}$ get infeasible.~Similarly, $a_{l,min}$ provides the minimum pruning rate for the current layer given all the next layers are pruned completely.~We clip the predicted action $a_l$ to lie in $[a_{l,min}, a_{l,max}]$ when pruning the $l$-th layer.

\section{Experimental Settings}
We provide more details of our experimental settings in the following.

\noindent\textbf{CIFAR-10:} For CIFAR-10 experiments, we evaluate our method on ResNet-56~\cite{he2016deep} and MobileNet-V2~\cite{sandler2018mobilenetv2}.~In our iterative pruning phase, we train both of the CNN models for $200$ epochs with the batch size of $128$ using SGD with momentum~\cite{sutskever2013importance} of $0.9$, weight decay of $1\mathrm{e}{-4}$, and starting learning rate of $0.1$.~We decay the learning rate by $0.1$ on epochs $100$ and $150$.~We take $5000$ samples of the training dataset as a subset for calculating the agent's reward.~For all cases, we start to train the RL agent after 10 warmup epochs of the model's weights.~Specifically, we collect initial data for the replay buffer of the RL agent from epochs 10 to 20.~Then, for both ResNet-56 and MobileNet-V2, we update the agent from epoch 20 until the epoch 90, and we train only the model's weights from epoch 90 to 200.~After the iterative stage, we prune the model's architecture and finetune it with the same settings for the base model.

\noindent\textbf{ImageNet:}~We use ResNet-18, ResNet-34, and MobileNet-V2 for ImageNet experiments.~For the iterative training stage of ResNets, we use SGD as the optimizer with the momentum of $0.9$, weight decay of $1\mathrm{e}{-4}$, and the start learning rate of $0.1$.~We train ResNet models for $90$ epochs, and we decay the learning rate to $0.01$ and $0.001$ at epochs $30$ and $60$.~For MobileNet-V2, we do so for $155$ epochs with a batch size of $256$.~We train the model's weights using SGD with the momentum of $0.9$, weight decay of $1\mathrm{e}{-4}$, and starting learning rate of $0.05$ decayed using cosine scheduling~\cite{loshchilov2017sgdr}.~For all cases, we use $50000$ samples of the training dataset to evaluate rewards of the agent.~Similar to CIFAR-10 experiments, we train the model's weights for 10 warmup epochs followed by 10 epochs for filling the replay buffer of the RL agent.~Then, for MobileNet-V2, we train the agent's policy from epochs 20 to 90, and we only train the model's weights from epoch 90 to 155.~After the pruning stage, we fine-tune the pruned model with the same training parameters as the base model. For ResNet models, we train the agent's policy from epochs 20 to 70, and the model's weights are trained from epochs 70 to 90.

\noindent\textbf{Ablation Experiments:} We follow the same settings as mentioned above for our ablation experiments in Tab.~3 of the paper. For the visualizations in Fig.~2, we use the same settings except that we perform our iterative pruning scheme for 90 epochs.

\end{document}